\documentclass[10pt,twocolumn,letterpaper]{article}

\usepackage{wacv}              

\usepackage[utf8]{inputenc} 
\usepackage[T1]{fontenc}    
\usepackage{hyperref}       
\usepackage{url}            
\usepackage{booktabs}       
\usepackage{amsfonts}       
\usepackage{nicefrac}       
\usepackage{microtype}      
\usepackage{xcolor}         
\usepackage{amsmath}

\usepackage{adjustbox}
\usepackage{graphicx}
\usepackage{enumitem}

\usepackage{xcolor}

\usepackage{mathtools}

\newcommand\scalemath[2]{\scalebox{#1}{\mbox{\ensuremath{\displaystyle #2}}}}

\usepackage[capitalize]{cleveref}
\crefname{section}{Sec.}{Secs.}
\Crefname{section}{Section}{Sections}
\Crefname{table}{Table}{Tables}
\crefname{table}{Tab.}{Tabs.}

\def\wacvPaperID{1520} 
\def\confName{WACV}
\def\confYear{2025}

\begin{document}

\title{Deep Geometric Moments Promote Shape Consistency in Text-to-3D Generation}



\author{
    Utkarsh Nath\textsuperscript{1}, Rajeev Goel\textsuperscript{1}, Eun Som Jeon\textsuperscript{2}\thanks{Corresponding Author}, Changhoon Kim\textsuperscript{1}, \\ Kyle Min\textsuperscript{3}, Yezhou Yang\textsuperscript{1}, Yingzhen Yang\textsuperscript{1}, Pavan Turaga\textsuperscript{1} \\
    \textsuperscript{1}Arizona State University\\
    \textsuperscript{2}Seoul National University of Science and Technology\\
    \textsuperscript{3}Intel Labs\\
}


\maketitle

\begin{abstract}
    
     To address the data scarcity associated with 3D assets, 2D-lifting techniques such as Score Distillation Sampling (SDS) have become a widely adopted practice in text-to-3D generation pipelines. However, the diffusion models used in these techniques are prone to viewpoint bias and thus lead to geometric inconsistencies such as the Janus problem. To counter this, we introduce MT3D, a text-to-3D generative model that leverages a high-fidelity 3D object to overcome viewpoint bias and explicitly infuse geometric understanding into the generation pipeline. Firstly, we employ depth maps derived from a high-quality 3D model as control signals to guarantee that the generated 2D images preserve the fundamental shape and structure, thereby reducing the inherent viewpoint bias. Next, we utilize deep geometric moments to ensure geometric consistency in the 3D representation explicitly. By incorporating geometric details from a 3D asset, MT3D enables the creation of diverse and geometrically consistent objects, thereby improving the quality and usability of our 3D representations. Project page and code: \url{https://moment-3d.github.io/}
\end{abstract}

\section{Introduction}
\label{sec:intro}

The creation of digital 3D content is essential in sectors like gaming, animation, and virtual/augmented reality. Traditionally, this process has been labor-intensive, requiring specialized expertise, which has spurred interest in automated text-driven 3D generation. Yet, text-to-3D generation faces major obstacles due to the lack of large-scale text-annotated 3D datasets. To overcome this, 2D lifting techniques ~\cite{chen2023fantasia3d, lin2023magic3d, metzer2023latent, poole2022dreamfusion, wang2023prolificdreamer} have emerged as a promising approach. These methods leverage pre-trained text-to-image diffusion models ~\cite{rombach2022high, nichol2021glide, saharia2022photorealistic} and use score distillation sampling (SDS) ~\cite{poole2022dreamfusion} to refine 3D models from textual inputs. Unlike traditional 3D supervised methods ~\cite{cheng2023sdfusion, jun2023shap, nichol2022point, wei2023taps3d}, SDS utilizes the extensive knowledge embedded in pre-trained text-to-image models. This approach bypasses the constraints of dataset scarcity and enhances the ability to generate innovative and previously unseen content.

\begin{figure}[t]
  \centering
  \includegraphics[width=1\linewidth]{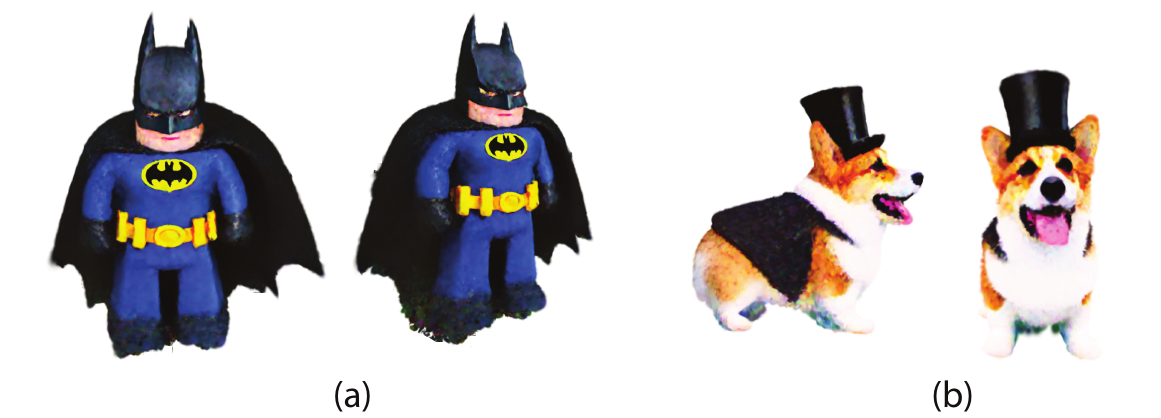}
   \vspace{-6pt}
   \caption{Illustration of 3D objects generated by our model corresponding to input text prompt (a) `A DSLR photo of Batman' and (b) `A high-quality photo of a corgi wearing a top hat'.}
   \label{fig:teaser}
   \vspace{-10pt}
\end{figure}

\begin{figure*}[t]
    \centering
    \vskip -0.2in
    \includegraphics[height=70pt]{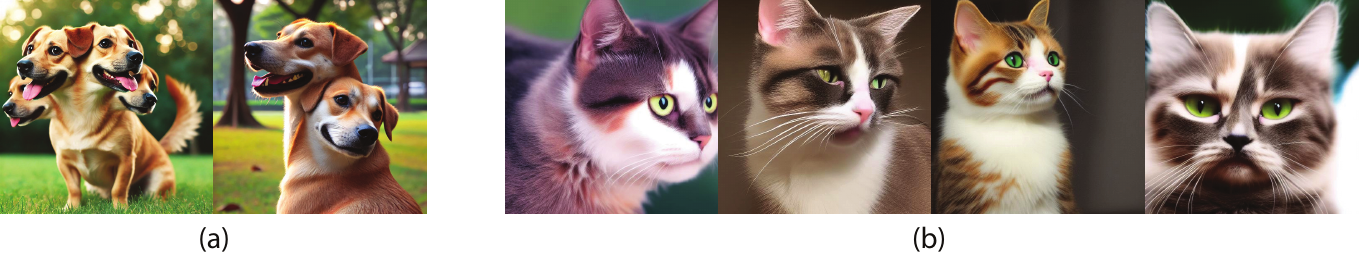}
        \caption{(a) An illustration of the multi-faced Janus problem. (b) An illustration of the inherent viewpoint bias in diffusion models. We present samples generated by Stable Diffusion v1.5 from the prompt 'cat,' where most samples predominantly depict the cat from a front view.}
        \label{fig:current_problem}
        \vspace{-10pt} 
\end{figure*}

However, commonly used 2D lifting techniques often suffer from geometric inconsistencies, such as the multi-faced Janus problem, vividly illustrated in Figure \ref{fig:current_problem}(a). This issue primarily stems from the inherent viewpoint biases in 2D diffusion models trained on extensive internet-scale datasets ~\cite{shi2023MVDream, li2023sweetdreamer, liu2023zero}, as demonstrated in Figure \ref{fig:current_problem}(b). These biases predispose the generated 3D models to overfit, typically manifesting as repetitive frontal views and other geometric distortions. 

Training diffusion models conditioned on camera parameters alleviates this problem. Liu \textit{et al.} ~\cite{liu2023zero} trained diffusion models on 2D images with their corresponding camera parameters. Despite incorporating knowledge from multiple views, the models remain confined to the 2D domain, lacking a comprehensive understanding of 3D geometry. Consequently, they continue encountering challenges with geometric inconsistencies when generating novel views. 

Nevertheless, 2D diffusion models are trained on large-scale datasets, which may be crucial for a diverse 3D generation. While we currently lack internet-scale 3D datasets necessary to directly train diffusion models for generating diverse 3D assets, we still have small/medium-sized high-fidelity 3D datasets ~\cite{deitke2023objaverse}. \textit{Can we use assets from high-fidelity 3D datasets to understand fundamental geometric structures while also taking advantage of the diverse representational power of pre-trained text-to-image models?} 

In this paper, we introduce MT3D, a novel moment-based text-to-3D generative model. MT3D is a 2D lifting technique designed to generate high-quality, geometrically consistent objects by learning geometry from a high-fidelity 3D object. Traditional 2D lifting techniques uniformly sample images from a 3D representation and subsequently align them with the high-probability images produced by the frozen diffusion model. Our objective is to reduce inherent viewpoint bias and incorporate 3D geometric understanding into our generation pipeline. 


Therefore, firstly, we utilize ControlNet ~\cite{zhang2023adding} to address viewpoint bias in diffusion models. ControlNet produces images based on its control signals. As depicted in Figure \ref{fig:controlnet_and_dgm}, images generated by ControlNet, when conditioned on various depth maps, demonstrate its ability to produce images from multiple viewpoints, thereby reducing viewpoint bias. In this paper, for each view sampled from the 3D representation, we render its corresponding view from a high-fidelity 3D object. We then extract a depth map from this rendered view, which serves as the control signal for the diffusion model. Along with ControlNet, we integrate a LoRA ~\cite{hu2022lora}, conditioned on the depth map, to enhance the diversity and quality of the outputs.

ControlNet mitigates viewpoint bias to a certain extent, but it performs poorly on views that are significantly scarce in the training dataset. For example, as shown in Figure \ref{fig:controlnet_and_dgm}, ControlNet generates a poor-quality image when conditioned on a depth map of a more challenging bottom view. In this paper, we employ deep geometric moments (DGM) ~\cite{nath2024polynomial, singh2023polynomial, SinghDLGC2023} to explicitly infuse the geometric properties of a high-fidelity 3D asset into the generated 3D representation, ensuring that even underrepresented views exhibit coherent geometry. Geometric moments define various shape characteristics and can be visualized as projections of the image onto chosen basis functions. Historically, the theoretical development of moments has often relied on specific selections of these bases ~\cite{teague1980image, hu1962visual}. Moments have been extensively studied in the vision community for a wide range of applications such as invariant pattern recognition ~\cite{hu1962visual, khotanzad1990invariant, alajlan2008geometry, flusser1993pattern}, segmentation ~\cite{reeves1988three, foulonneau2006affine}, and 3D shape recognition ~\cite{tuceryan1994moment,sadjadi1980three, elad2003bending}. 

 In our approach, we employ a DGM network pre-trained on ImageNet to extract higher-order geometric moment-like features. The DGM is inspired by traditional moment calculations, with the significant difference that its basis functions are learned end-to-end through deep learning. DGM effectively generates rich and discriminative features that capture the inherent shape and geometry of objects (Figure \ref{fig:controlnet_and_dgm}). We propose a loss function that minimizes the difference between the DGM features obtained by the estimated 3D representation and those of the high-fidelity 3D object.

 Explicit incorporation of geometric knowledge facilitates the generation of diverse and geometrically coherent 3D representations. In this paper, we utilize 3D Gaussians to represent our 3D assets. Compared to other 3D representation such as NeRF ~\cite{mildenhall2021nerf, muller2022instant, barron2021mip}, Gaussian Splatting ~\cite{kerbl20233d} offers two distinct advantages. Firstly, the geometry of 3D Gaussians can be initialized using point cloud priors, which is not feasible with NeRF due to its implicit nature. Secondly, rendering time for 3D Gaussians is significantly lower than NeRF.

Our contributions can be summarized as:
\begin{enumerate}[leftmargin=*]
    \item We present MT3D, a 3D Gaussian-based 2D lifting technique that incorporates geometric knowledge directly from a high-fidelity 3D object into our generative process. 

    \item We propose employing ControlNet, conditional-LoRA and a DGM-based loss function to mitigate viewpoint bias and reduce inconsistencies in 3D generation.
    
    \item We validate our methods using various text prompts and against various state-of-the-art generators ~\cite{lin2023magic3d, chen2023fantasia3d, zhu2023hifa, wang2023prolificdreamer, chen2023text}. Empirical evaluations suggest that MT3D can generate well-structured and high-fidelity 3D objects. It achieves 38\% better Janus rate than the next most effective generator, HiFA ~\cite{zhu2023hifa}. 
\end{enumerate}




\section{Related Work}
\label{sec:related_work}
 \textbf{Text-to-3D Generation} creates 3D images from textual descriptions. This field is divided into two main approaches: 3D supervised and 2D lifting techniques. 3D supervised techniques ~\cite{cheng2023sdfusion, jun2023shap, nichol2022point, wei2023taps3d} utilize paired text and 3D data for training, efficiently producing detailed 3D models. However, their generalizability is limited because of the small sizes of available 3D datasets. In contrast, 2D lifting techniques ~\cite{poole2022dreamfusion, lin2023magic3d, chen2023fantasia3d, wang2023prolificdreamer, zhu2023hifa, huang2023dreamcontrol, chen2023text} use pre-trained 2D diffusion models learned on internet-scale datasets to refine 3D representations. For instance, DreamFusion ~\cite{poole2022dreamfusion} introduced score distillation sampling (SDS) to refine 3D content by aligning rendered images with a 2D diffusion prior. Fantasia3D ~\cite{chen2023fantasia3d} separates the processes for learning geometry and materials, using physics-based rendering for precise mesh generation. ProlificDreamer ~\cite{wang2023prolificdreamer} improves SDS with variational score distillation to generate diverse, high-quality 3D assets. HiFA ~\cite{zhu2023hifa} applies SDS across both latent and image dimensions of the diffusion model. While these techniques can produce photorealistic images, they often struggle with geometric inconsistencies, such as the Janus problem. To overcome these issues, recent methods ~\cite{li2023sweetdreamer, liu2023zero, liu2023syncdreamer, long2023wonder3d, seo2023let, shi2023MVDream} utilize small-scale 3D datasets to train 3D priors, integrating these into their models to enhance diffusion processes. However, training on limited datasets can reduce the generalizability of models. Additionally, the generation process frequently leads to a loss of high-fidelity texture, which can be traced back to the cartoonish style of the 3D data utilized. DreamControl ~\cite{huang2023dreamcontrol} addresses this by using a coarse 3D model as a self-prior for better consistency. In our research, we utilize a high-quality 3D object to learn geometric features and structures, without relying on any additional datasets. Similar to methods in ~\cite{chen2023text}, our approach includes geometric and appearance refinement stages. Unlike previous techniques, our pipeline specifically utilizes a high-quality 3D object and geometric moments to effectively infuse geometric knowledge into our model, significantly improving the geometric consistency of our 3D representation.

\textbf{Controllable Generation.}
Text inputs are widely used as primary control inputs for 3D generative tasks. However, alternative control signals such as images ~\cite{ruiz2023dreambooth, wei2023elite}, videos ~\cite{zhang2023controlvideo, chen2023control}, and sketches ~\cite{metzer2023latent} also play a crucial role in guiding the generation process. ControlNet ~\cite{zhang2023adding} enhances text-to-image synthesis by integrating additional conditions like edges, depth maps, and normals. This method supports the creation of 3D content directed by 2D sketches ~\cite{chen2023control3d} and video sequences ~\cite{shao2023control4d}. In this paper, we employ a depth-conditioned ControlNet to ensure geometric consistency in the generated 3D objects. Depth conditions serve as effective proxies for defining 3D spatial information.

\vspace{-10pt}
\section{Preliminaries}

\textbf{Gaussian Splatting} ~\cite{kerbl20233d} 
models a scene with a collection of 3D Gaussians, characterized by means, covariances, colors, and opacities. These are transformed into 2D Gaussians on the image plane for rendering, with color integration performed via alpha blending in a front-to-back sequence. Compared to NeRF-based approaches ~\cite{barron2022mip, muller2022instant}, Gaussian Splatting achieves finer reconstructions, enhances rendering speeds, and reduces memory demands during training. Rendering for Gaussian Splatting is expressed as $r=g(\theta, c)$, where $\theta$ represents the 3D Gaussians, $g$ is the renderer, $c$ indicates the camera parameters and $r$ is the rendered image.

\textbf{Score Distillation Sampling (SDS)} ~\cite{poole2022dreamfusion} 
optimizes a 3D representation $\theta$ using a pre-trained 2D diffusion model $\phi$. The 3D representation is optimized such that images rendered from any camera projection closely resemble those generated by the pre-trained 2D text-to-image diffusion model. The SDS gradient is formulated as,
\begin{equation}
\nabla_{\theta} \mathcal{L}_{\text{SDS}}(\theta) = \mathbb{E}_{t,\epsilon} \left[ \omega(t) \left( \hat{\epsilon}_{\phi}(r_t, t, y) - \epsilon \right) \frac{\partial r}{\partial \theta} \right],
\label{eqn:sds}
\end{equation}
where, $\hat{\epsilon}_{\phi}$ represents the noise predicted by the model $\phi$, and $r_t$, $t$, and $y$ correspond to the noisy image, timestep, and text embedding, respectively. $\epsilon$ denotes Gaussian noise, and $\omega(t)$ is the weighting function applied at each timestep.


\textbf{Geometric Moment} of a two-dimensional, piece-wise continuous function \( f(x, y) \) is defined as follows:
\begin{equation}
\label{eq:momentdef}
m_{pq} = \int_{-\infty}^{\infty} \int_{-\infty}^{\infty} x^{p} y^{q} f(x, y) \, dx \, dy,
\end{equation}
where \((x, y)\) represent the 2D coordinates, and \(p+q\) denotes the order of the moment. Geometric moments can be conceptualized as a ‘projection’ of the 2D function onto polynomial bases of the form \(x^p y^q\). Image moments, recognized as invariant shape descriptors, have been extensively utilized in computer vision for analyzing and interpreting the geometrical attributes of an image. Pioneering work by Hu ~\cite{hu1962visual} introduced seven lower-order invariant moments that remain unchanged under scaling, translation, and rotation. Subsequent studies ~\cite{chong2004translation, zhang2011affine, zhang2009blurred, khotanzad1990invariant, kim2003invariant, wang1998using, yap2005efficient, flusser2003moment} have expanded these concepts, introducing invariant moments for arbitrary orders.

\begin{figure*}[t]
    \centering
    \includegraphics[width=0.9\textwidth]{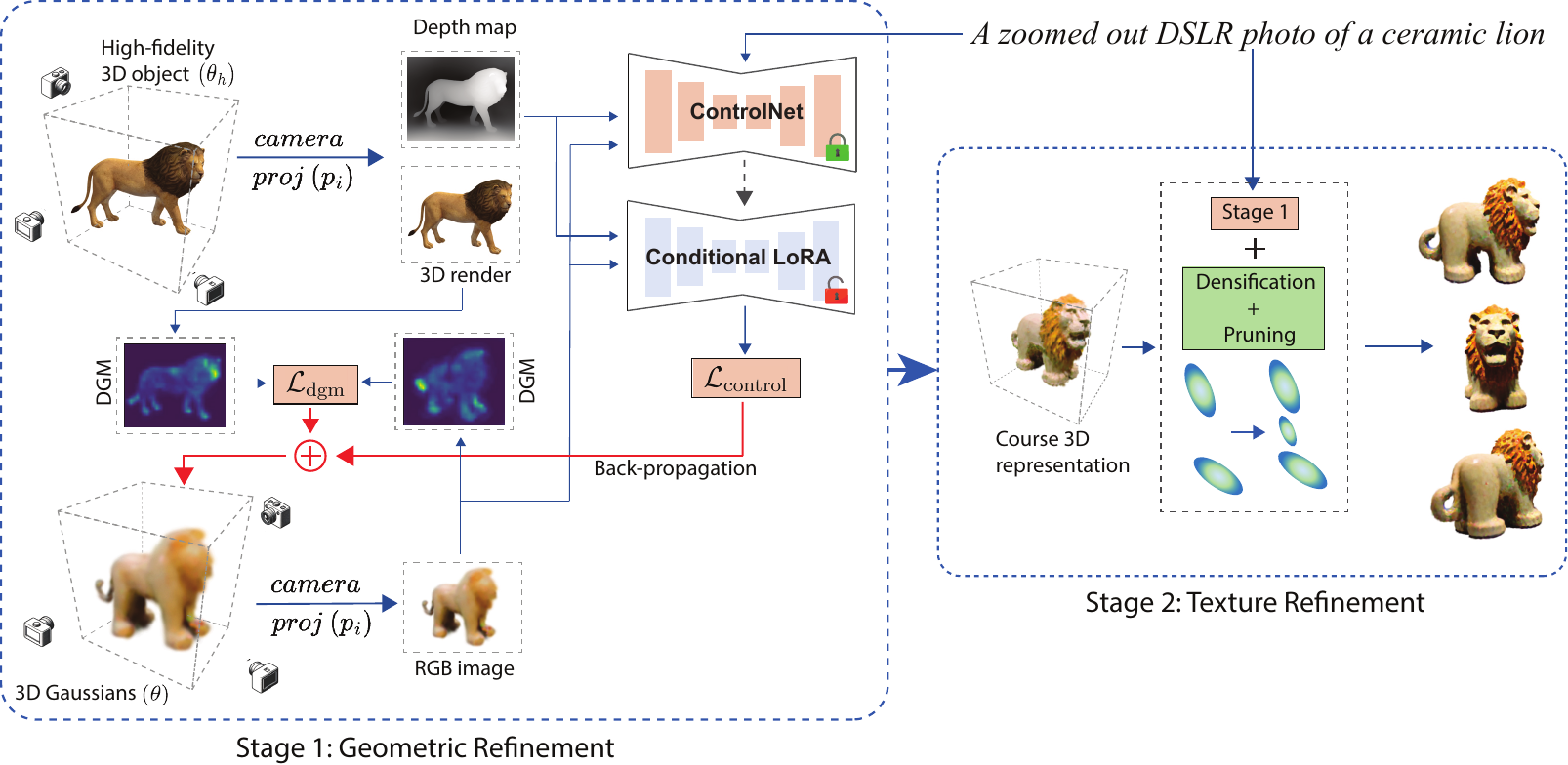}
        \caption{
    Overview of MT3D. In the first stage, we optimize 3D Gaussians using a high-fidelity 3D object with depth-conditioned ControlNet and deep geometric moments (DGM). Red and green locks represent frozen and trainable weights, respectively. The second stage extends the first by not only utilizing ControlNet and DGM features but also applying additional densification and pruning.}
    \vspace{-10pt}
        \label{fig:mt3d}
\end{figure*}

\section{Methods}
\label{sec:method}

Our objective is to develop high-fidelity, diverse, and geometrically consistent 3D objects from textual descriptions. While existing 2D-to-3D lifting methods can create novel and diverse scenarios based on text inputs, they often face issues with geometric inconsistencies, such as the multi-faced Janus problem, due to a lack of explicit 3D geometric knowledge. In this paper, we aim to close this gap by embedding explicit geometric information into our model using a high-quality 3D object, thus improving the realism of our generated 3D objects. Specifically, upon receiving a text prompt, we identify a suitable high-fidelity 3D object (Section \ref{subsection:3d_object}), and utilize this representation to enhance our model's geometric understanding.
%

We adopt 3D Gaussians for our 3D representations, valuing their flexibility in incorporating geometric priors and their efficiency in rendering. Following  ~\cite{chen2023fantasia3d, tang2023make, huang2023dreamcontrol, chen2023text}, our methodology also includes a dual-phase training strategy. The first phase focuses on refining the geometric aspects of the 3D representation (Section \ref{subsection:geometry}), and the subsequent phase concentrates on enhancing the textural details (Section \ref{subsection:appearence_refinement}).

\begin{figure*}[t]
    \centering
    \includegraphics[width=0.98\textwidth]{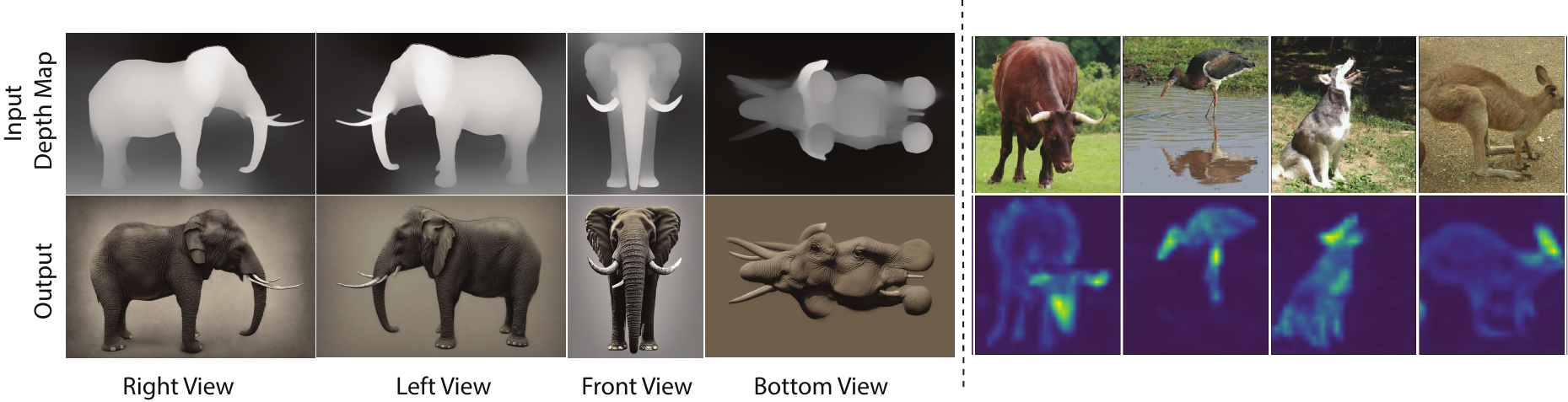}
    \vskip -0.1in
        \caption{(\emph{Left}) Samples generated by the depth-conditioned ControlNet model from the prompt `Elephant,' conditioned on various viewpoints. The outputs from the ControlNet model align well with the majority of the input-conditioned depth maps. (\emph{Right}) Illustration of features obtained from the ImageNet-pretrained deep geometric moment Model (DGM) across various images. DGM features effectively capture the shape and structure.}
        \label{fig:controlnet_and_dgm}
\end{figure*}

\subsection{Retrieving high-fidelity 3D object}
\label{subsection:3d_object}
Upon receiving a text prompt, we select a high-quality 3D object from Objaverse ~\cite{deitke2023objaverse}, a moderately-sized dataset of diverse 3D assets, to steer the generation process. For every object within Objaverse, we create a corresponding caption using Cap3D ~\cite{luo2024scalable}. These captions, along with the input text prompt, are then processed through SBert ~\cite{reimers-2019-sentence-bert} to calculate cosine similarities. The object that shows the highest similarity to the text prompt is chosen for the next steps in the generation sequence. It is important to note that our generation pipeline is compatible with any 3D representation. In this section, we have automated the selection of high-fidelity 3D objects for ease of implementation.

\subsection{Geometric Refinement}
\label{subsection:geometry}
In SDS optimization, we aim to reduce the distribution discrepancy between images rendered from the 3D model and those generated by a pre-trained 2D diffusion model. However, a significant challenge arises from the viewpoint bias inherent in these diffusion models ~\cite{shi2023MVDream, li2023sweetdreamer, liu2023zero}, which may hinder the model's ability to generate diverse views corresponding to different camera settings. This often leads to uniform appearances across various camera angles, contributing to the Janus problem (Figure \ref{fig:current_problem}(a)). GSGEN ~\cite{chen2023text} addressed these geometric inconsistencies by incorporating a 3D SDS loss alongside a pre-trained point cloud diffusion model. They found that 3D Gaussians could be effectively adjusted using point cloud priors, a strategy not viable with NeRFs due to their implicit nature. Therefore, to enhance the 3D generation process, in addition to the standard SDS, they also employ Point-E ~\cite{nichol2022point}, a text-to-point cloud diffusion model, guided by the following equation:
\begin{multline}
\nabla_{\theta} \mathcal{L}(\theta) = \lambda_{3D} \cdot \mathbb{E}_{t,\epsilon_{p}} \left[ \omega_{p}(t) \left( \hat{\epsilon}_{\psi}(p_t, t, y_e) - \epsilon_{p} \right) \right] + \\
\mathbb{E}_{t,\epsilon} \left[ \omega(t) \left( \hat{\epsilon}_{\phi}(r_t, t, y_e) - \epsilon \right) \frac{\partial r}{\partial \theta}  \right],
\label{eqn:gsgen}
\end{multline}
where $p_t$ is noisy Gaussian position, and $\omega_{*}$ and $\epsilon_{*}$ refer to the corresponding weighting function and Gaussian noise, respectively. Applying SDS with Point-E generally results in better geometry compared to standard SDS. However, they still suffer from geometric inconsistencies due to two primary factors. Firstly, the 2D diffusion model suffers from viewpoint bias. Secondly, there is no explicit optimization focused on the overall shape or structure of the 3D representation. We address these challenges in Section \ref{subsection:control_3d} and \ref{subsection:dgm}.

\vspace{-10pt}
\subsubsection{Controlling 3D Generation}
\label{subsection:control_3d}

In this section, we aim to address the Janus problem associated with SDS by mitigating viewpoint bias. Additionally, SDS demonstrates limited diversity as it treats 3D representations as a single point in real space. To overcome these challenges, we incorporate ControlNet ~\cite{zhang2023adding} and Variational Score Distillation (VSD) ~\cite{wang2023prolificdreamer}.


\textbf{Utilizing ControlNet for mitigating viewpoint bias.} In our approach, we utilize a high-fidelity 3D object ($\theta_{h}$) as a control signal for ControlNet. Figure \ref{fig:controlnet_and_dgm} demonstrates that ControlNet, when provided with a conditional image, can generate images from a variety of viewpoints, effectively reducing the viewpoint bias associated with standard diffusion models. During SDS, for each view rendered from the 3D representation ($\theta$), we also project a corresponding view from $\theta_{h}$ using identical camera parameters. We use the depth map of the rendered view as a guiding signal for ControlNet, ensuring that the generated 2D image preserves the essential shape and structure.

\textbf{Improving diversity using VSD.} VSD treats the 3D representation as a distribution range instead of a single point, as seen in SDS. This approach enhances both the quality and diversity of the outputs. VSD utilizes a particle-based update method through the Wasserstein gradient flow. The VSD gradient is formulated as, 
\begin{equation}
\scalemath{0.89}{
\nabla_{\theta} \mathcal{L}_{\text{VSD}}(\theta) = \mathbb{E}_{t,\epsilon} \left[ \omega(t) \left( \hat{\epsilon}_{\phi}(r_t, t, y) - 
\hat{\epsilon}_{\theta}(r_t, t, c, y)
\right) \frac{\partial r}{\partial \theta} \right],
}
\label{eqn:vsd}
\end{equation}
where $\hat{\epsilon}_{\theta}(r_t, t, c, y)$ is the noise predicted by rendered images. In practice, $\epsilon_{\theta}$ is parameterized by a LoRA ~\cite{hu2022lora} of the pre-trained model conditioned with camera parameter $c$ and optimized by, 
\begin{equation}
\mathbb{E}_{t \sim \mathcal{U}(0,1), \epsilon \sim \mathcal{N}(0, I)} \left[ \left\| \hat{\epsilon}_\theta(\alpha_t x_t + \sigma_t \epsilon, t, c, y) - \epsilon \right\|^2_2 \right].
\label{eqn:lora}
\end{equation}


In our approach, we replace SDS with VSD to broaden the range of generation for a given prompt. However, optimizing LoRA with explicit conditioning on camera parameters proves challenging. Therefore, we refine the model to implicitly incorporate camera parameters by substituting $c$ with depth maps ($\theta_{hc}^{d}$). Consequently, LoRA is conditioned on depth maps derived from the high-fidelity images ($\theta_{h}$) corresponding to the camera parameters $c$. The training objective for LoRA is the same as Eq. \eqref{eqn:lora}. However, initially, LoRA lacks practical significance and tends to predict random noise. To address this, we initially set the weights of the LoRA term to zero and gradually increase them over iterations using a cosine schedule. As a result, $\hat{\epsilon}_{\phi}(r_t, t, y) - 
\hat{\epsilon}_{\theta}(r_t, t, c, y)$ term in Eq. \eqref{eqn:vsd} is replaced by 
\begin{equation}
  \hat{\epsilon}_{\phi}(r_t, t, \theta_{hc}^{d}, y) - 
\lambda_{lora}.\hat{\epsilon}_{\theta}(r_t, t, \theta_{hc}^{d}, y),  
\end{equation}
where $\hat{\epsilon}_{\phi}(r_t, t, \theta_{hc}^{d}, y)$ is the noise predicted by depth-conditioned ControlNet and $\lambda_{lora}$ is a weighting term.

In summary, as illustrated in Figure \ref{fig:mt3d}, at each optimization step, we render an RGB image $r$ from $\theta$ and a depth map ($\theta_{hc}^{d}$) from $\theta_{h}$ using the same camera parameters $c$. The depth map serves as a conditional image for ControlNet and LoRA. Thus, the optimization step in Eq. \eqref{eqn:gsgen} is revised as follows:
\begin{equation}
 \resizebox{0.9\hsize}{!}{$
 \begin{split}
\nabla_{\theta} \mathcal{L}_{\text{control}}(\theta) = \lambda_{p} \cdot \mathbb{E}_{t,\epsilon_{p}} \left[ \omega_{p}(t) \left( \hat{\epsilon}_{\psi}(p_t, t, y_e) - \epsilon_{p} \right) \right] + \\
\mathbb{E}_{t,\epsilon} \left[ \omega(t) \left( \hat{\epsilon}_{\phi}(r_t, t, \theta_{hc}^{d}, y_e) - \lambda_{lora}.\hat{\epsilon}_{\theta}(r_t, t, \theta_{hc}^{d}, y) \right) \frac{\partial r}{\partial \theta}  \right].
 \end{split}
 $}
 \label{eqn:loss_control}
\end{equation}

Following ~\cite{wang2023prolificdreamer}, we alternatively update gradients of Eq. \eqref{eqn:loss_control} and \eqref{eqn:lora}.

\subsubsection{Deep Geometric Moment}
\label{subsection:dgm}
In the previous section, we address the viewpoint bias prevalent in diffusion models. ControlNet effectively mitigates this bias to a degree, but it performs poorly on views that are significantly underrepresented in the training data. 
For instance, as illustrated in Figure \ref{fig:controlnet_and_dgm}, while ControlNet successfully generates left, right, and front views, it fails to produce the more challenging bottom view.
In this section, we utilize geometric moments to explicitly induce the geometric properties of a high-fidelity 3D object into the generated 3D representation, ensuring that even underrepresented views maintain coherent geometry.

\textbf{Promoting shape awareness.} Classical geometric moments are highly effective at capturing shape information and providing discriminative cues. However, their discriminative power is relatively limited and typically requires the presence of a salient object against a homogeneous background. In response to this limitation, recent work ~\cite{nath2024polynomial, singh2023polynomial, SinghDLGC2023} has introduced deep geometric moments (DGM). This approach trains a neural network to learn higher-order geometric moments in an end-to-end differentiable manner. DGM excels at capturing the geometric shapes and structures within 2D images. Figure \ref{fig:controlnet_and_dgm} displays features extracted using the ImageNet pre-trained DGM model. In this paper, we minimize the gap between the DGM features obtained from our 3D representations and those of the high-fidelity 3D object to explicitly promote shape awareness.

One could envision other ways to promote shape controls, including edge-map comparisons or semantic segmentation map comparisons. Edge-maps are a lower-level representation and are sensitive to the background, textures, color variations, etc. Further, comparing two edge-maps is often done by specialized metrics such as Chamfer matching ~\cite{borgefors1988hierarchical}. Semantic segmentation masks, on the other hand, are a higher-level representation but can suffer from over- or under-segmentation. Matching two segmentation maps also requires specialized metrics. We find that the deep geometric moment (DGM) occupies a unique intermediate space, where it highlights important parts of the object, particularly the outlines in a more robust and invariant way than edge-maps, yet is lower-level than segmentation masks, and can be treated simply as a feature that can be compared using the standard L-2 loss. We adopt the DGM representation here due to these favorable and convenient properties.


\vspace{-10pt}
\subsubsection{Overall Optimization}
Finally, for the comprehensive geometric refinement of the 3D representation, we employ both the depth-conditioned ControlNet and DGM module. We define the DGM loss for a randomly chosen 2D projection , $r$, from the 3D representation ($\theta$) as:
$\mathcal{L}_{\text{dgm}}(\theta) = ||DGM(r) -  DGM(\theta_{hc}) ||_{2}$, where $DGM(r)$ represents the features obtained from a frozen Imagenet pre-trained DGM model for 2D projection $r$, $\theta_{hc}$ is the corresponding high-fidelity image projection and $\|\cdot\|_2$ represents the $\ell^2$-norm. Thus, the overall optimization process can be formulated as follows:
\begin{equation}
    \nabla_{\theta} \mathcal{L}_{\text{geometry}}(\theta) = \mathcal{L}_{\text{control}}(\theta) +
    \lambda_{m}.\mathcal{L}_{\text{dgm}}(\theta),   \label{eqn:geometry_refinement}
\end{equation} 
where, $\lambda_{m}$ is a regularization constant.

\begin{figure*}[t]
    \centering
    \includegraphics[width=0.95\textwidth]{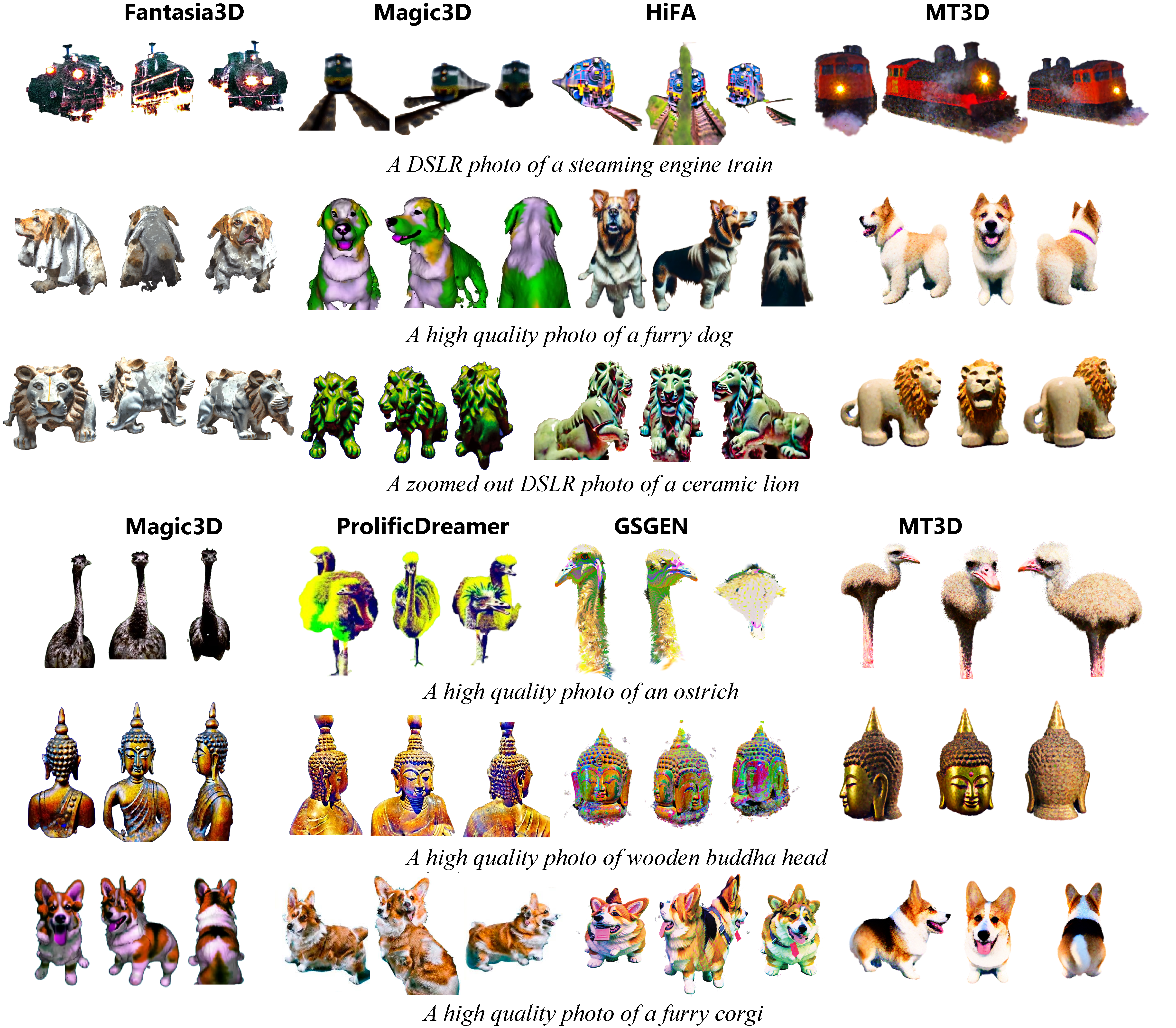}
        \caption{Qualitative comparison between the proposed MT3D and state-of-the-art generators, including Magic3D ~\cite{lin2023magic3d}, Fantasia3D ~\cite{chen2023fantasia3d},  ProlificDreamer ~\cite{wang2023prolificdreamer}, HIFA ~\cite{zhu2023hifa} and GSGEN ~\cite{chen2023text}.}
        \label{fig:Comparison_sota}
        \vspace{-10pt}
\end{figure*}

\subsection{Texture Refinement}
\label{subsection:appearence_refinement}
The texture refinement stage builds upon geometric refinement, where we progressively refine and increase the density of the 3D Gaussians. Following the approach detailed in ~\cite{kerbl20233d}, our strategy targets Gaussians for splitting based on their significant view-space spatial gradients. We also incorporate a strategy for densification based on compactness, as suggested by ~\cite{chen2023text}, by inserting new Gaussians where the spatial gap between existing ones surpasses their combined radii; the radius of the new Gaussian is adjusted to bridge the remaining gap. Moreover, we periodically remove Gaussians, considering their opacity and size. The loss function applied in this process is identical to geometric refinement (Eq. \eqref{eqn:geometry_refinement}).

\section{Experiment}
\label{sec:exp}
We utilize the publicly available Stable Diffusion v1-5 ~\cite{poole2022dreamfusion} and depth-conditioned ControlNet ~\cite{zhang2023adding} as our pre-trained diffusion models. Adopting the DreamFusion methodology, we employ view-dependent prompts while ensuring consistent focal length, elevation, and azimuth ranges. We implement our 3D Gaussian Splatting rendering pipeline using PyTorch. We train both the geometric and appearance refinement stages for 15,000 iterations. Generating a 3D asset typically requires approximately four hours on an A100 machine. For texture refinement densification, we adopt the method described in ~\cite{chen2023text}, where we split Gaussians based on the view-space position gradient at every 500 iterations with a threshold of 0.02, and perform compactness-based densification every 1,000 iterations. We prune Gaussians with an opacity below 0.05 or a radius greater than 0.05 every 500 iterations. We set $\lambda_p$ in Eq. \eqref{eqn:loss_control} to 1.0 and $\lambda_{m}$ in Eq. \eqref{eqn:geometry_refinement} to 100 to ensure all terms of the loss function are on a similar scale. During the geometric refinement stage, we adjust $\lambda_{lora}$ following a cosine schedule, incrementing it from 0 to 0.75 across the initial 5,000 steps. Subsequently, we maintain this value constant for the remaining training period and throughout the texture refinement stage.

\subsection{Text-to-3D Generation}
\label{subsection:comparison_with_sota}
 We assess MT3D's performance in the text-to-3D generation domain through both qualitative and quantitative comparisons against various state-of-the-art methods ~\cite{lin2023magic3d, chen2023fantasia3d, zhu2023hifa, wang2023prolificdreamer, chen2023text}. Another notable work, Dreamcontrol ~\cite{huang2023dreamcontrol}, also achieves state-of-the-art results; however, we are unable to compare with them as their code is not available at the time of writing this paper. As depicted in Figures \ref{fig:Comparison_sota} and \ref{fig:teaser}, MT3D consistently produces high-quality, well-structured 3D assets with consistent shape and structure. In contrast, alternative approaches often demonstrate significant geometric inconsistencies when responding to the same text prompts. For instance, with the ceramic lion prompt, both Magic3D and Fantasia3D create assets with multiple faces. When prompted with the dog, HiFA generates images with extra legs, and Fantasia3D's model outputs a dog with disoriented legs and body. Conversely, MT3D consistently delivers accurate and precise geometric renderings of both lions and dogs. Similarly, for challenging prompts like `train' and `ostrich,' where methods such as ProlificDreamer, GSGen, and HiFA fall short in producing photorealistic images and often struggle with multiple facial features, MT3D excels by generating assets that are not only photorealistic but also accurately depict shape and structure. 

 We also quantitatively evaluate MT3D against other state-of-the-art (SOTA) methods. To assess geometric consistency, we measure the occurrence rate of the Janus problem, defined under conditions such as: (1) the presence of multiple faces, hands, legs, or similar anomalies; (2) noticeable content drift; (3) significant occurrences of paper-thin structures. The Janus Rate is calculated by dividing the number of instances with inconsistent content by the total number of generated objects. Table \ref{tab:jr} presents the Janus rates for various leading 3D generators. The results underscore MT3D’s superior performance in maintaining consistent 3D geometries, with a Janus rate of 38\% better than the next most effective generator, HiFA.
 
 \begin{table}[h]
\centering
\caption{Illustration of Janus Rate for various state-of-the-art 3D generators, evaluated across twenty diverse text prompts. $\downarrow$ signifies that a lower value is preferable.}
\label{tab:jr}
\begin{tabular}{|c|c|}
\hline
\textbf{Method} & \textbf{Janus Rate $\downarrow$} \\ \hline
Magic3D ~\cite{lin2023magic3d} &   75    \\ \hline
HiFA ~\cite{zhu2023hifa} &   58    \\ \hline
GSGEN ~\cite{chen2023text} &   63   \\ \hline
MT3D (Ours) &  36    \\ \hline
\end{tabular}
\vspace{-15pt}
\end{table}

\begin{figure*}[t]
    \centering
    \includegraphics[width=0.90\textwidth]{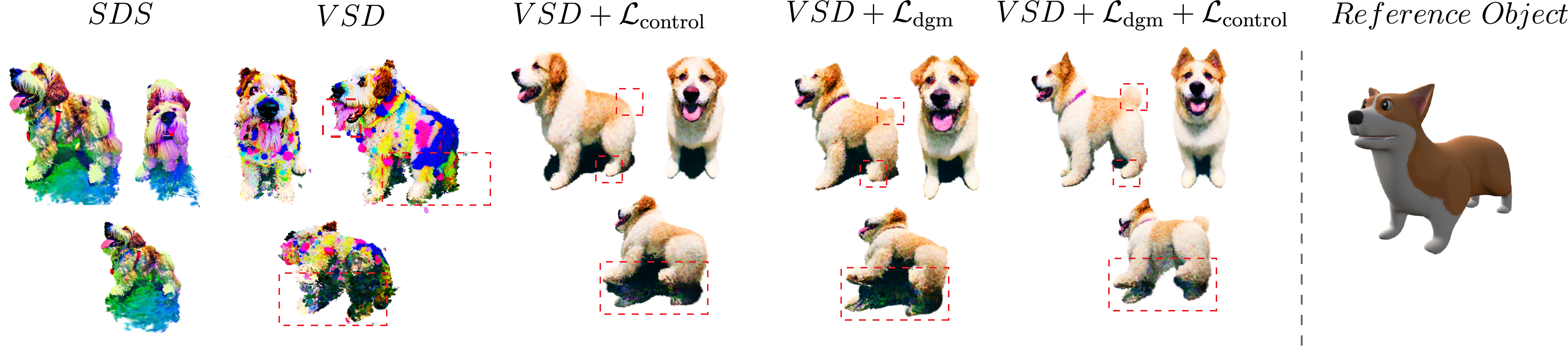}
    \vskip -0.07in
        \caption{Results of an ablation study on MT3D with the text prompt: "A high-quality photo of a furry dog." The reference object represents a high-fidelity 3D model used to guide MT3D’s generation process. Additional ablation studies can be found in Figure \ref{fig:more_ablation} of the appendix.}
        \label{fig:ablation}
\end{figure*}

\subsection{Ablation Study}
\label{subsection:ablation}
In this section, we present an ablation study to evaluate the impact of various components in our 3D generation pipeline, specifically focusing on controlling the 3D generation, deep geometric moment and the use of a high-fidelity 3D object. Figure \ref{fig:ablation} showcases the comparison among five generative pipelines. The `SDS' configuration refers to models optimized using Eq. \ref{eqn:gsgen} during Stage 1. `VSD' represents the model optimized by using VSD (\ref{eqn:vsd}) instead of SDS (\ref{eqn:sds}) in Eq. \ref{eqn:gsgen}. `VSD + $\mathcal{L}_{\text{control}}$' indicates models trained with the additional Eq. \ref{eqn:loss_control}. `VSD + $\mathcal{L}_{\text{moment}}$' signifies models trained with both Eq. \ref{eqn:gsgen} and the DGM loss ($\mathcal{L}_{\text{dgm}}$). Lastly, `VSD + $\mathcal{L}_{\text{control}}$ + $\mathcal{L}_{\text{moment}}$' describes our proposed pipeline, as detailed in Eq. \ref{eqn:geometry_refinement}.

Models optimized with the SDS and VSD configuration exhibit various geometric inconsistencies, including multiple faces, indistinct legs, and random noise at the bottom. However, integrating each component of MT3D leads to significant enhancements in overall shape and structure. Although $\mathcal{L}_{\text{dgm}}$ and $\mathcal{L}_{\text{control}}$ appear similar at first glance, notable differences emerge upon closer inspection. For example, in the generated representation of a `furry dog', the side view of the dog's leg is better shaped under $\mathcal{L}_{\text{control}}$ than under $\mathcal{L}_{\text{dgm}}$. Conversely, the bottom view is more accurately rendered with $\mathcal{L}_{\text{dgm}}$ than with $\mathcal{L}_{\text{control}}$. This can be attributed to the likelihood that ControlNet's training dataset predominantly comprises side views of dogs, rather than bottom views, resulting in a better generation of side views compared to bottom views. Thus, ControlNet focuses more on learning the general aspects of shape, while DGM primarily learns shape and structure from the reference object. By combining $\mathcal{L}_{\text{dgm}}$ and $\mathcal{L}_{\text{control}}$, MT3D delivers superior fidelity across all perspectives. Additional ablation studies in Section \ref{subsection:more_ablation} confirm consistent results across various text prompts.




\begin{figure}[!h]
    \centering
    \includegraphics[width=0.95\linewidth]{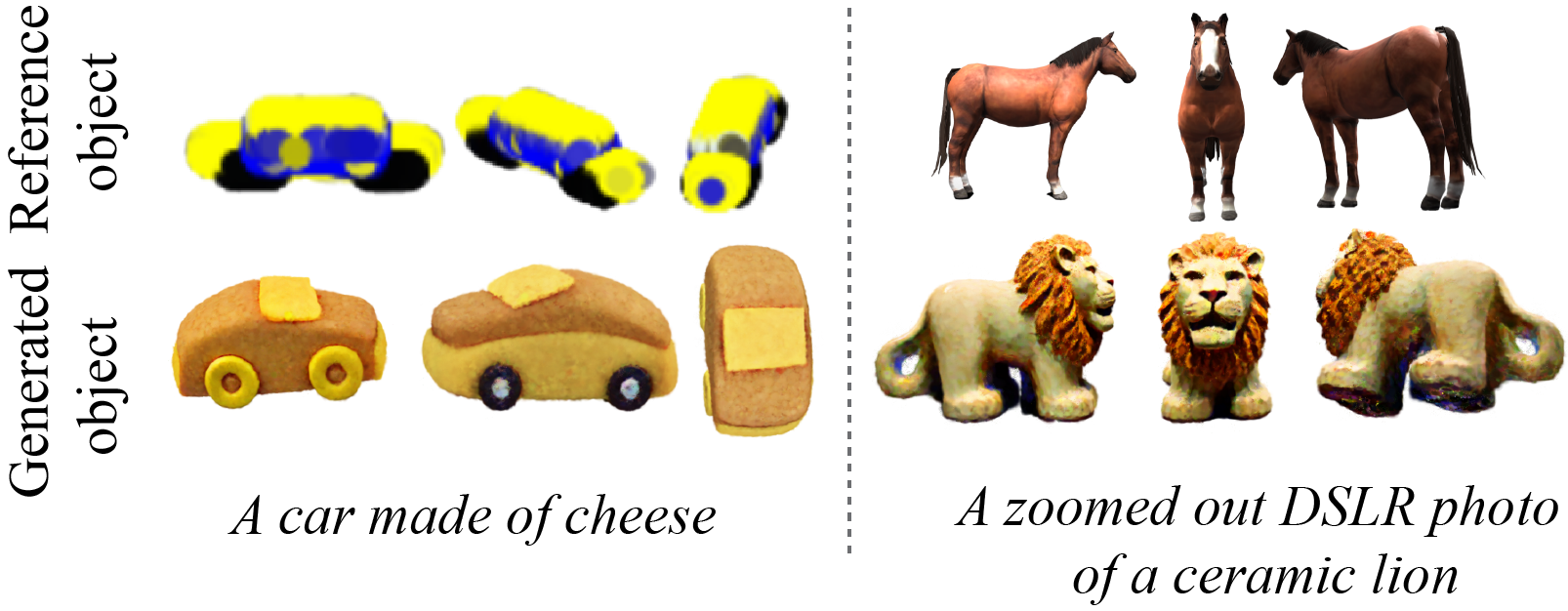}
    \vskip -0.07in
        \caption{(left) 3D objects generated by MT3D using low-fidelity reference objects from Point-E. (Right) 3D objects generated by MT3D using analogous reference objects. Please refer to Section \ref{subsection:unavailable_reference_obj} of the appendix for more illustrations.}
        \label{fig:not_reference_obj}
\end{figure}

\textbf{Absence of high-fidelity reference 3D object: } MT3D relies on reference objects for geometrically coherent 3D representations. We evaluate its performance using two alternative guidance types in the absence of high-fidelity references. First, we use low-fidelity or geometrically inconsistent 3D representations generated by other text-to-3D models, such as Point-E ~\cite{nichol2022point}. Second, instead of using a high-fidelity 3D object that directly matches the text prompt, we explore using 3D objects from a similar category for guidance. For example, in response to a `ceramic lion' prompt, we use a 3D model of a horse as guidance. As shown in Figure \ref{fig:not_reference_obj}, our model demonstrates the ability to generate well-structured and high-fidelity 3D representations, even when the geometric guidance is limited or geometrically inconsistent. We conduct exhaustive experiments with various types of reference objects in Section \ref{subsection:unavailable_reference_obj} of the appendix. 

\section{Conclusion}
\label{sec:conclusion}
In this study, we present MT3D, a cutting-edge 2D-lifting technique for text-to-3D generation that targets the geometric inconsistencies found in current methodologies by leveraging a high-fidelity 3D object. Our approach integrates depth-conditioned ControlNet, LoRA, and deep geometric moments, significantly improving the geometric coherence and fidelity of the produced 3D objects. Comprehensive experiments and ablation studies confirm the efficacy of MT3D in addressing challenges like the Janus problem and enhancing the overall structure and shape. While our primary focus has been on generating well-defined shapes and structures, the texture of the generated 3D objects remains critically important. Future work may explore whether texture information can also be learned from high-fidelity 3D objects.

\section*{Acknowledgements}
This work was supported in part by NSF grant 2323086, and by a Subcontract from The Johns Hopkins University with funds provided by an Other Transaction Agreement, No. HR00112490422 from The Defense Advance Research Project Agency.

{\small
\bibliographystyle{ieee_fullname}
\bibliography{references}
}

\newpage
\appendix
\newpage
\section{Appendix} 

\subsection{Unavailability of Reference Objects}
\label{subsection:unavailable_reference_obj}
MT3D relies on a high-fidelity 3D object to generate geometrically coherent representations. In this section, we assess the performance of MT3D in situations where an object that directly matches the text prompt is unavailable. We propose two alternative approaches for these scenarios. Firstly, we suggest utilizing existing text-to-3D generators to create initial 3D representations, which can then guide our generation pipeline. Secondly, instead of selecting a high-fidelity object that precisely matches the text prompt, we investigate the feasibility of using 3D objects that approximately belong to the same class as the object of interest.

\subsubsection{Utilizing Existing Text-to-3D Generators}
\label{subsubsection:using_text_to_3d}
\begin{figure*}[t]
    \centering
    \includegraphics[width=0.95\textwidth]{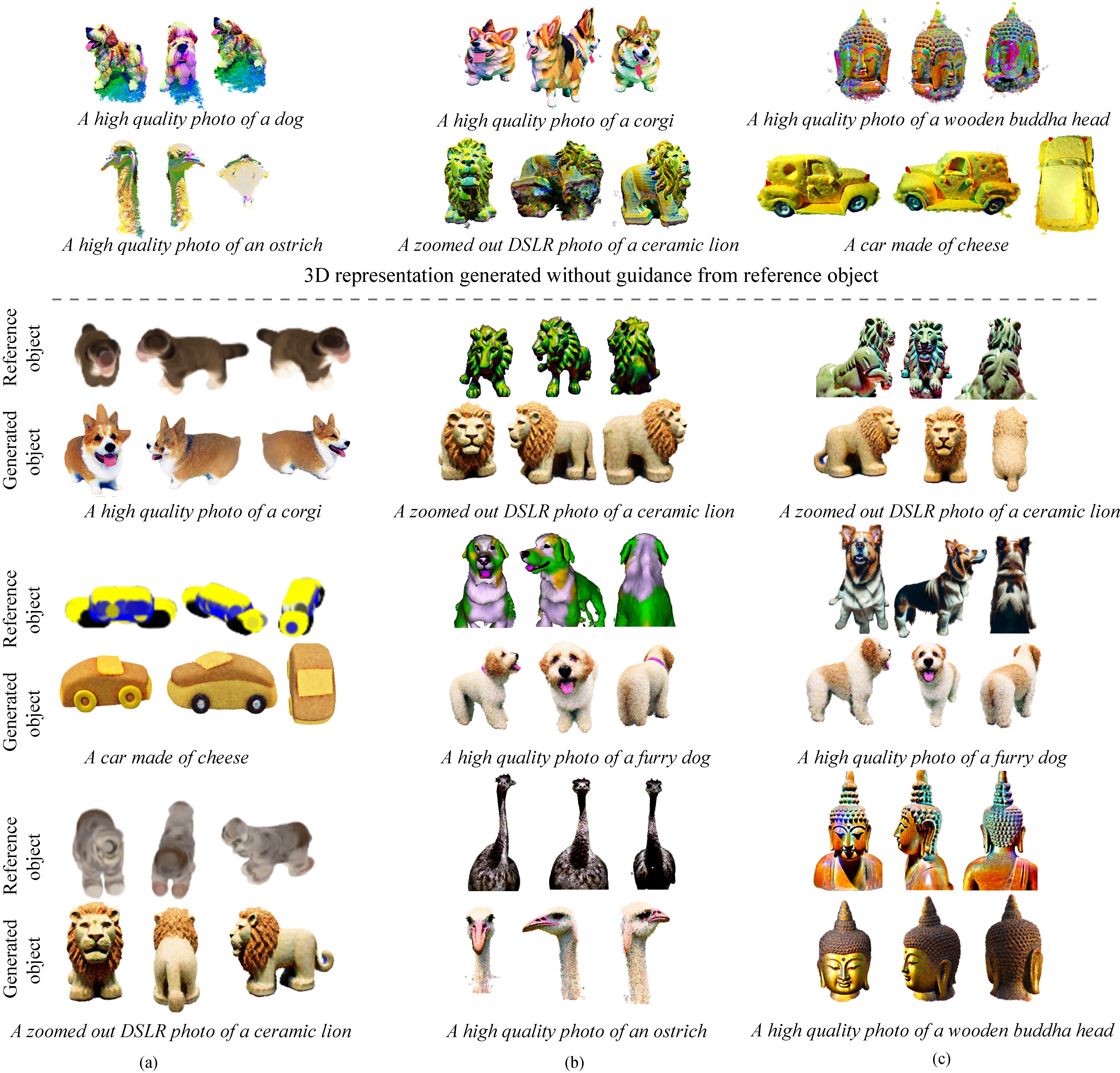}
        \caption{(Top) Illustration of 3D objects generated by SDS, as described in Figure \ref{fig:ablation}). (Bottom) Illustration of 3D objects generated by MT3D using various reference objects for different input text prompts. The 3D reference objects are generated from (a) Point-E, (b) Magic3D, and (c) HiFA.}
        \label{fig:different_3d_reference}
\end{figure*}

To assess the generative capabilities of MT3D, we evaluate its performance using objects generated by existing text-to-3D generators as guiding references. Specifically, we experiment with three types of reference objects: 1) low-fidelity objects generated by Point-E ~\cite{nichol2022point}, 2) high-fidelity objects generated by HiFA, and 3) objects exhibiting the Janus problem. Objects generated by Point-E typically exhibit low fidelity and limited diversity due to its direct training on 3D datasets. Moreover, Point-E often produces objects with poor shape and structure when handling complex prompts involving out-of-distribution objects. However, even when guided by such low-fidelity reference objects with disoriented geometry and poor structural quality, MT3D successfully generates 3D structures with minimal geometric disorientations. For example, as shown in Figure \ref{fig:different_3d_reference}(a), MT3D generates a high-fidelity ceramic lion, preserving a coherent shape and structure despite the deficiencies of the reference object. Although certain generated representations, such as the corgi, display the Janus problem, these outputs demonstrate notable improvement compared to those generated without any reference guidance.

The quality of MT3D generation improves further when it is guided by more geometrically coherent reference objects. For example in Figure \ref{fig:different_3d_reference} (c), when using objects generated by HiFA as references, MT3D produces high-fidelity 3D models with consistent geometry. Even when the reference object has minor geometric inconsistencies, such as the lion and corgi generated by HiFA, MT3D still generates representations with a coherent shape and structure. This robustness can be attributed to the reliance on geometric moments, which focus on the overall shape and structure while being invariant to minor local changes. Additionally, to comprehensively evaluate MT3D’s capabilities, we experiment with reference objects exhibiting geometric inconsistencies and the Janus problem (Figure \ref{fig:different_3d_reference} (b)). In these cases, MT3D produces representations with significantly improved shape and structure compared to the reference objects. It is important to note that MT3D leverages both a text-based 2D image generator i.e ControlNet and deep geometric moments (DGM) for generation. Consequently, even when the reference object is of poor quality and the features learned by the DGM module are suboptimal, the text-based ControlNet generator attempts to compensate for these shortcomings and generate a decent 3D representation.

\begin{figure*}[t]
    \centering
    \includegraphics[width=0.90\textwidth]{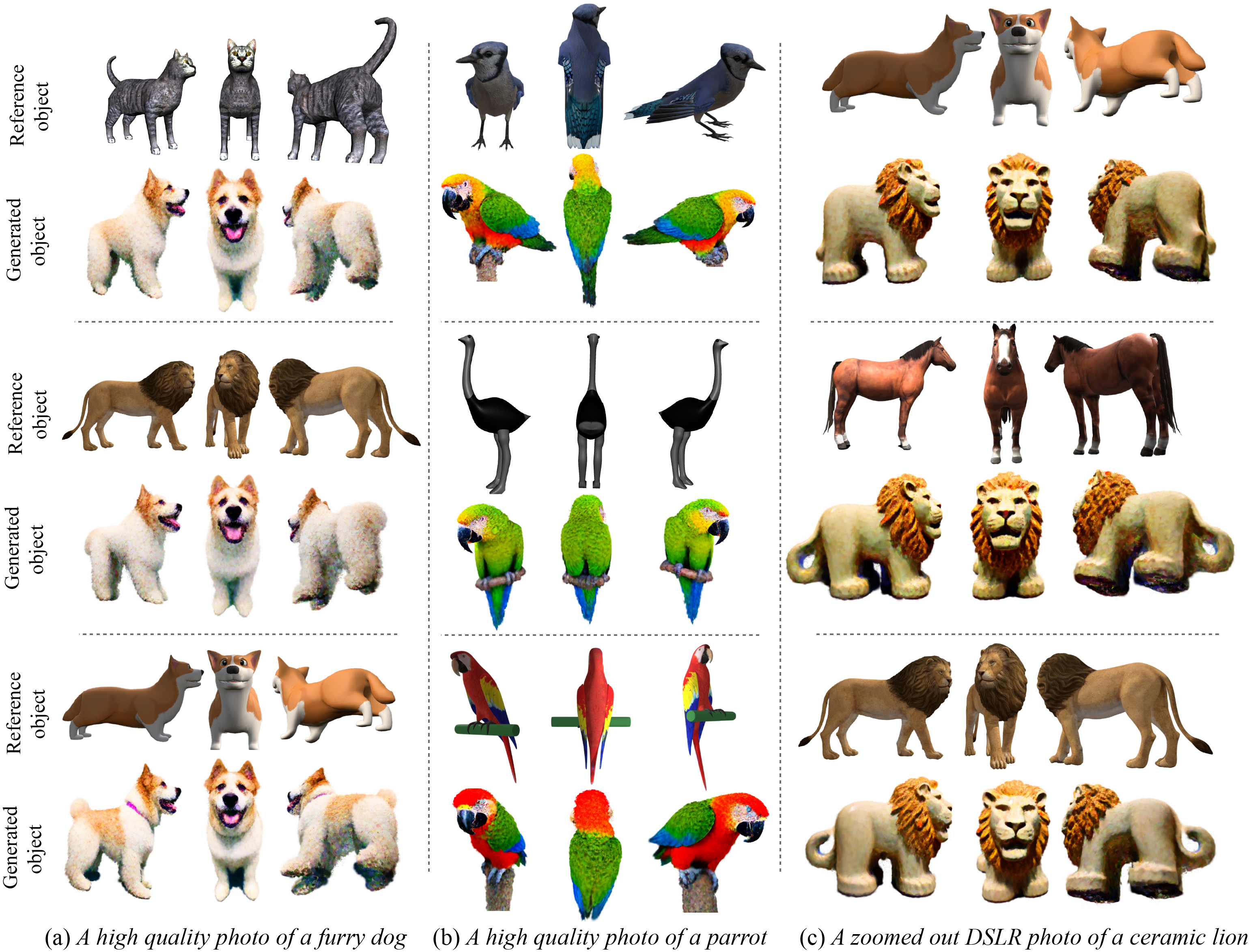}
        \caption{Illustration of 3D objects generated by our model using various high-fidelity objects corresponding to different input text prompts. The top row in each block represents the input high-fidelity object, and the bottom row shows the generated 3D asset.}
        \label{fig:analogous_final}
\end{figure*}

\subsubsection{Analogous geometric guidance}
\label{subsubsection:analogous}
Instead of using a high-fidelity 3D reference object that directly matches the text prompt, we explore using 3D objects from a similar category for guidance. For example, in response to a "furry dog" prompt, we provide guidance using a 3D representation of a cat or a lion, rather than a dog. As shown in Figure \ref{fig:analogous_final}, despite using objects from analogous categories, MT3D successfully generates well-structured and high-fidelity 3D representations. This result stems from the fact that geometric moments capture global shape and structure, which are not significantly affected by fine-grained details. This underscores MT3D’s ability to generalize and effectively utilize geometric cues from similar objects, even in the absence of an exact match.

\subsection{More Ablation Study}
\label{subsection:more_ablation}

\begin{figure*}[t]
    \centering
    \includegraphics[width=0.95\textwidth]{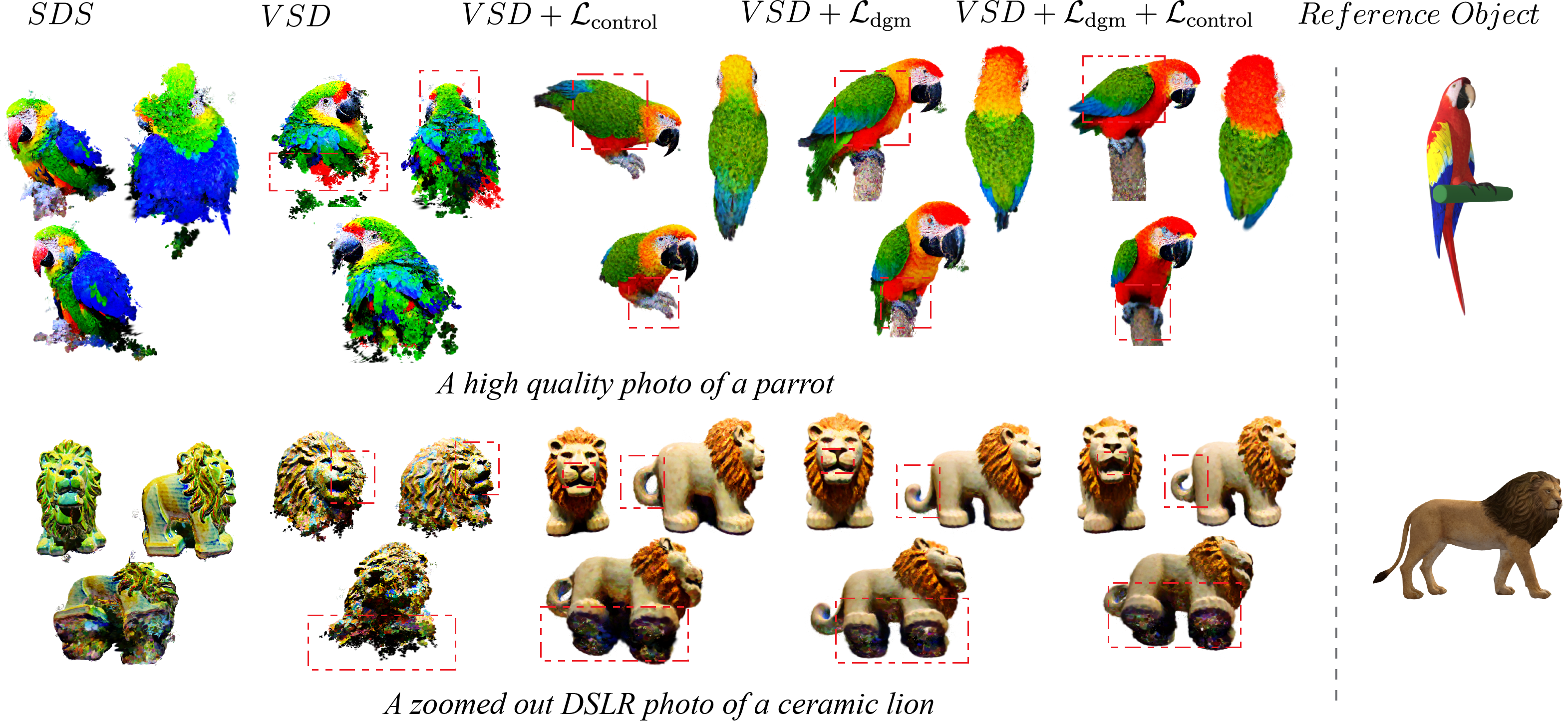}
        \caption{The results of an ablation study of MT3D.}
        \label{fig:more_ablation}
\end{figure*}

In this section, we present additional ablation studies (Figure \ref{fig:more_ablation}). Consistent with Section \ref{subsection:ablation}, models optimized with the SDS and VSD configurations exhibit various geometric inconsistencies. Similarly, the bottom view and tail of the lion are better shaped under $\mathcal{L}_{\text{dgm}}$, while the face of the lion exhibits more detailed features with $\mathcal{L}_{\text{control}}$. Furthermore, the effect of DGM is evident in the case of the parrot, where $\mathcal{L}_{\text{dgm}}$ successfully generates a branch similar to the reference object. In contrast, ControlNet is unable to do so, likely because its training data may not have included many instances of parrots perched on a branch. Thus, ControlNet focuses more on learning the general aspects of shape, while DGM primarily learns shape and structure from the reference object. By leveraging both $\mathcal{L}_{\text{dgm}}$ and $\mathcal{L}_{\text{control}}$, MT3D produces superior representations across the front, back, side, and bottom views, thereby enhancing the overall fidelity of shape and structure. 

\subsection{Comparison against Fantasia3D, Magic3D and HiFA}
\label{subsection:comparison_with_3_sota}


\begin{figure}[t]
    \centering
    \includegraphics[width=0.95\columnwidth]{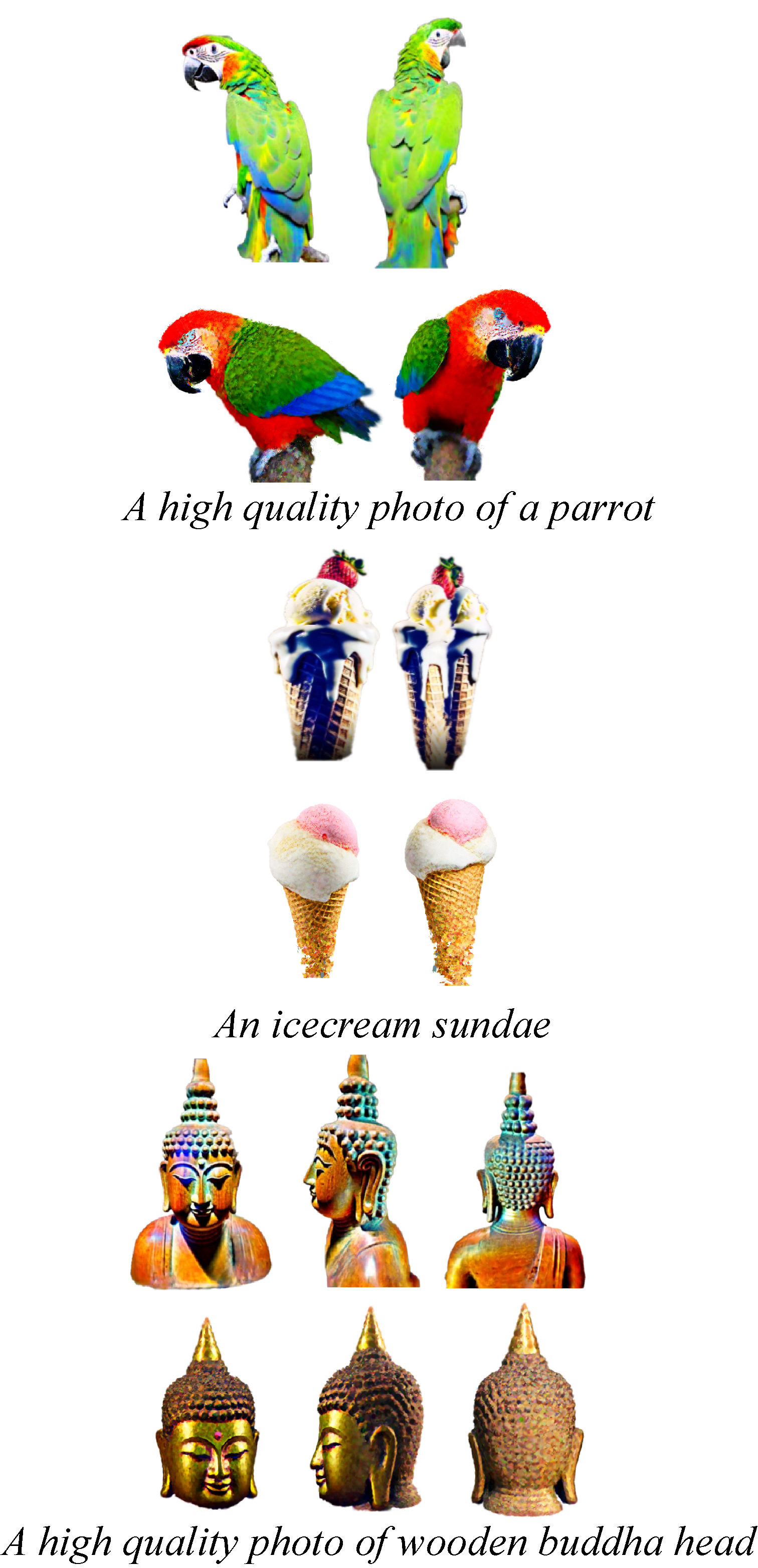}
        \caption{Qualitative comparison of 3D assets generated by MT3D and HiFA. The top row for each prompt displays HiFA-generated objects, while the bottom row shows MT3D-generated assets.}
        \label{fig:hifa121}
\end{figure}

\begin{figure*}[t]
    \centering
    \includegraphics[width=0.83\linewidth]{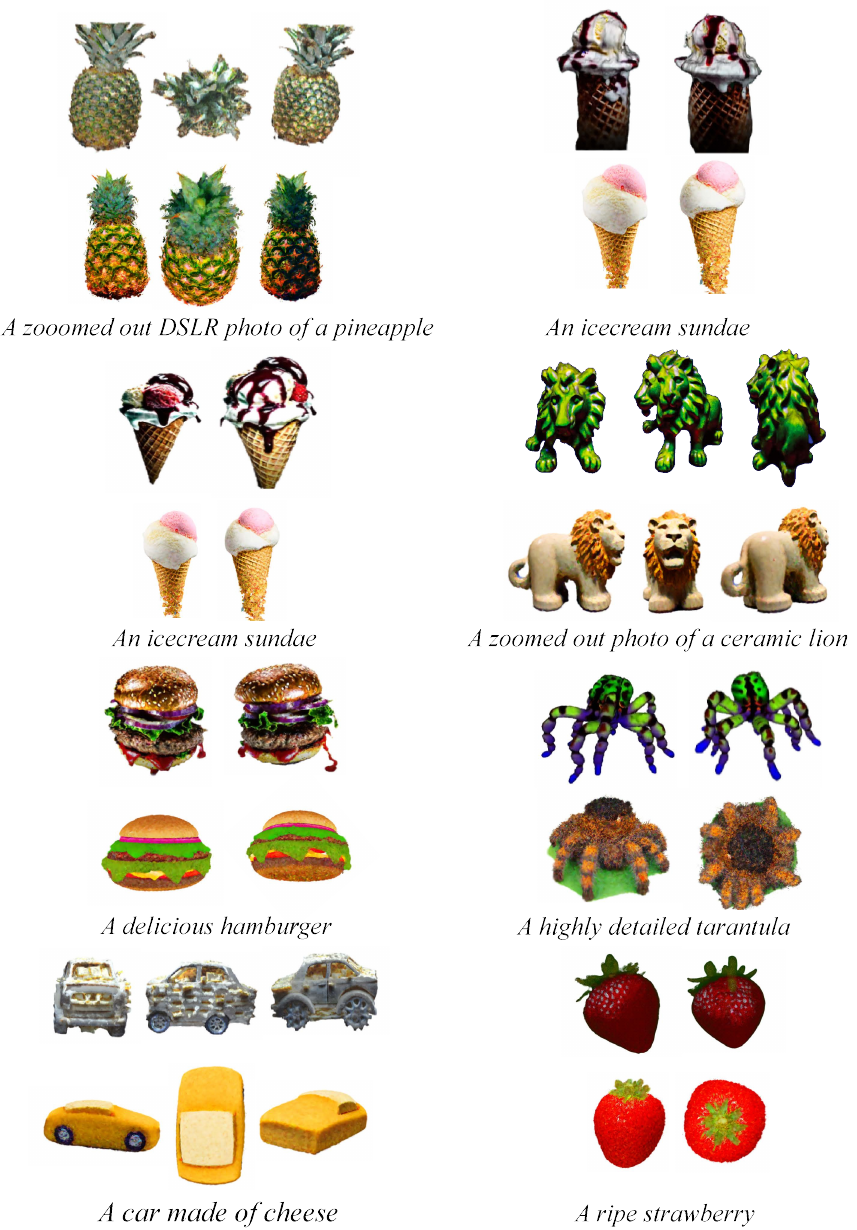}
        \caption{(Left) Qualitative comparison of 3D assets generated by MT3D and Fantasia3D. (Right) Qualitative comparison of 3D assets generated by MT3D and Magic3D. For each prompt, the top row shows assets from Fantasia3D (Left) and Magic3D (Right), while the bottom row displays those generated by MT3D.}
        \label{fig:fantasia121}
\end{figure*}


In Figure \ref{fig:hifa121} and \ref{fig:fantasia121}, we qualitatively compare MT3D with state-of-the-art methods—Fantasia3D, Magic3D, and HiFA—using text prompts from their respective original papers. For Fantasia3D and HiFA, we utilize their original source code to generate 3D representations. Since the original code for Magic3D was unavailable at the time of writing, we use the threestudio-based ~\cite{threestudio2023} implementation to generate its representations. All experiments were conducted on 4 $\times$ A100 GPUs with a batch size of 32 and random seeds. We observed that the 3D representations generated by these state-of-the-art methods closely match those reported in the original papers. Across all prompts, MT3D consistently produces high-fidelity and geometrically coherent 3D representations.

\subsection{Complex Prompts}
\label{subsection:complex_prompt}
\begin{figure*}[h]
    \centering
    \includegraphics[width=1.0\linewidth]{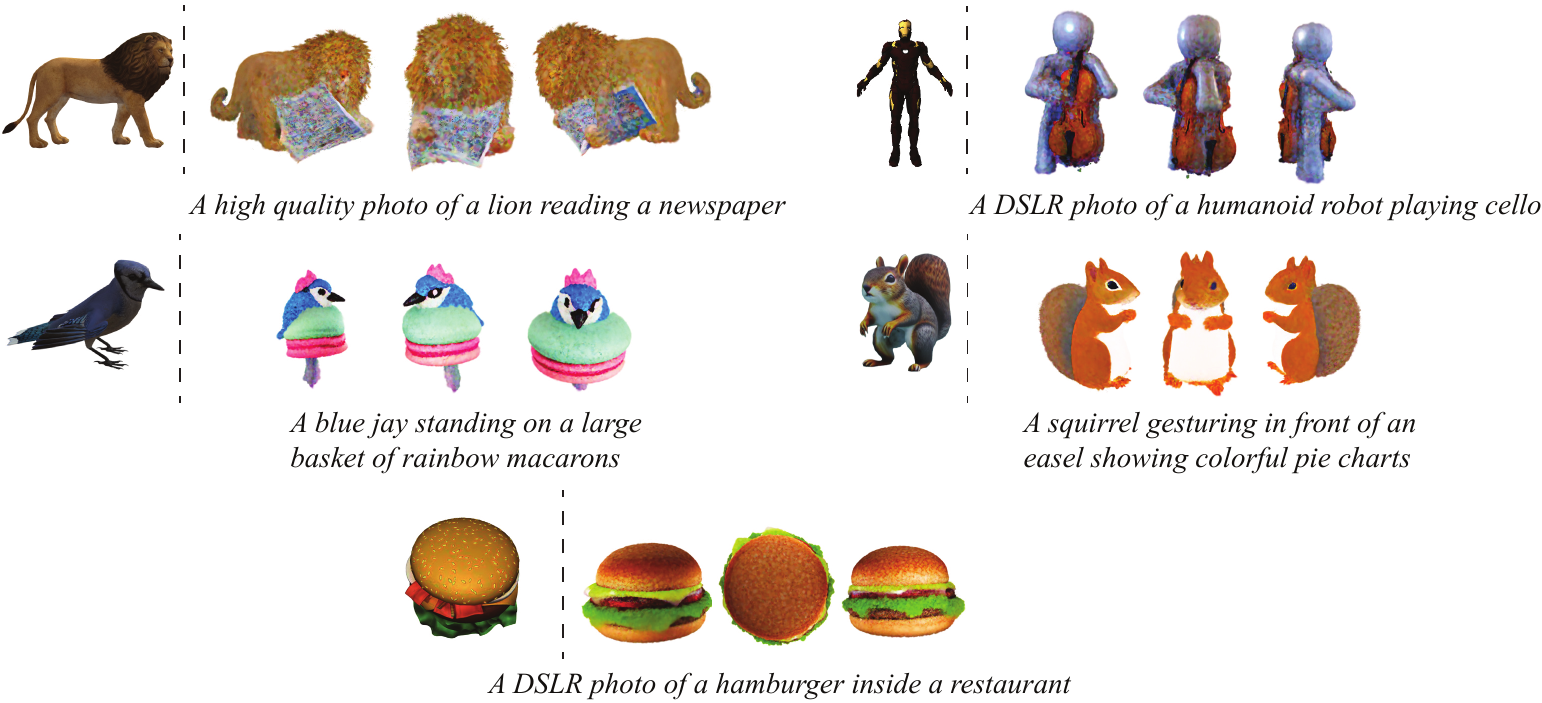}
        \caption{Illustrations of 3D objects generated by MT3D using various complex prompts. For each text prompt, the left shows the high-fidelity reference object, while the right represents the 3D asset generated by MT3D using the reference object.}
        \label{fig:complex_prompt}
\end{figure*}

In Figure \ref{fig:complex_prompt}, we conduct experiments with more complex text prompts including multiple objects and in different scenes. MT3D utilizes a single 3D representation to guide the generation process, which limits its ability to generate multiple objects or place objects within a particular scene. Across all complex prompts, the generated 3D representations exhibit a bias toward the geometry of the reference object. For example, in text prompt corresponding to `blue jay standing on a large basket of rainbow macarons', MT3D generates a blue jay and a macaron but no basket. Similarly, for `hamburger in restaurant', no restaurant was generated. Utilizing multiple reference objects may help address this issue. Extending MT3D to incorporate multiple reference objects, thereby enabling the generation of multiple objects and complex scenes, would be a valuable direction for future work.

\subsection{Additional Results}
\begin{figure*}[h]
    \centering
    \includegraphics[width=0.95\linewidth]{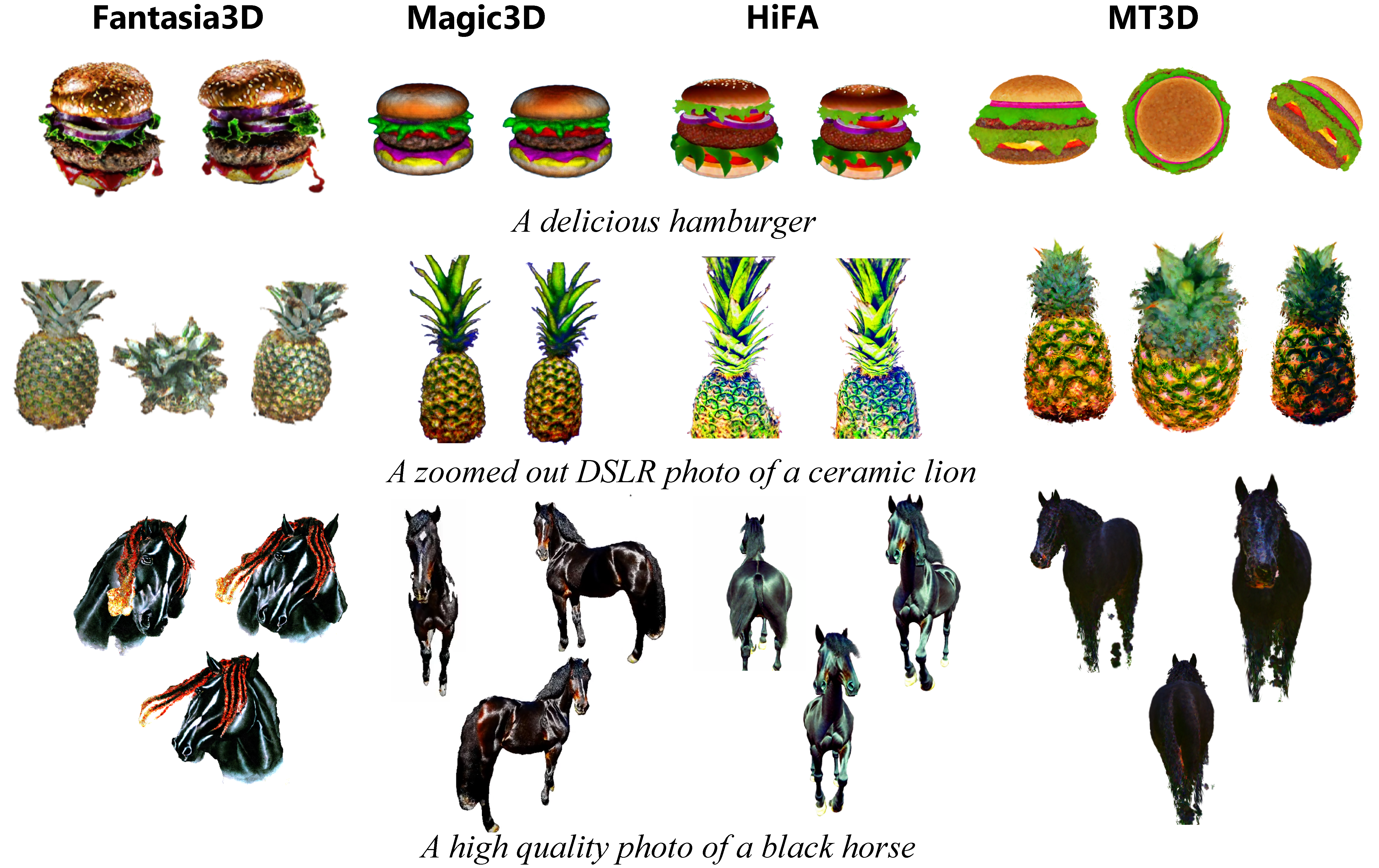}
        \caption{Additional qualitative comparison between the proposed MT3D and state-of-the-art generators, including Magic3D, Fantasia3D, and HiFA}
        \label{fig:Comparison_sota2}
\end{figure*}

\begin{figure*}[h]
    \centering
    \includegraphics[width=0.95\linewidth]{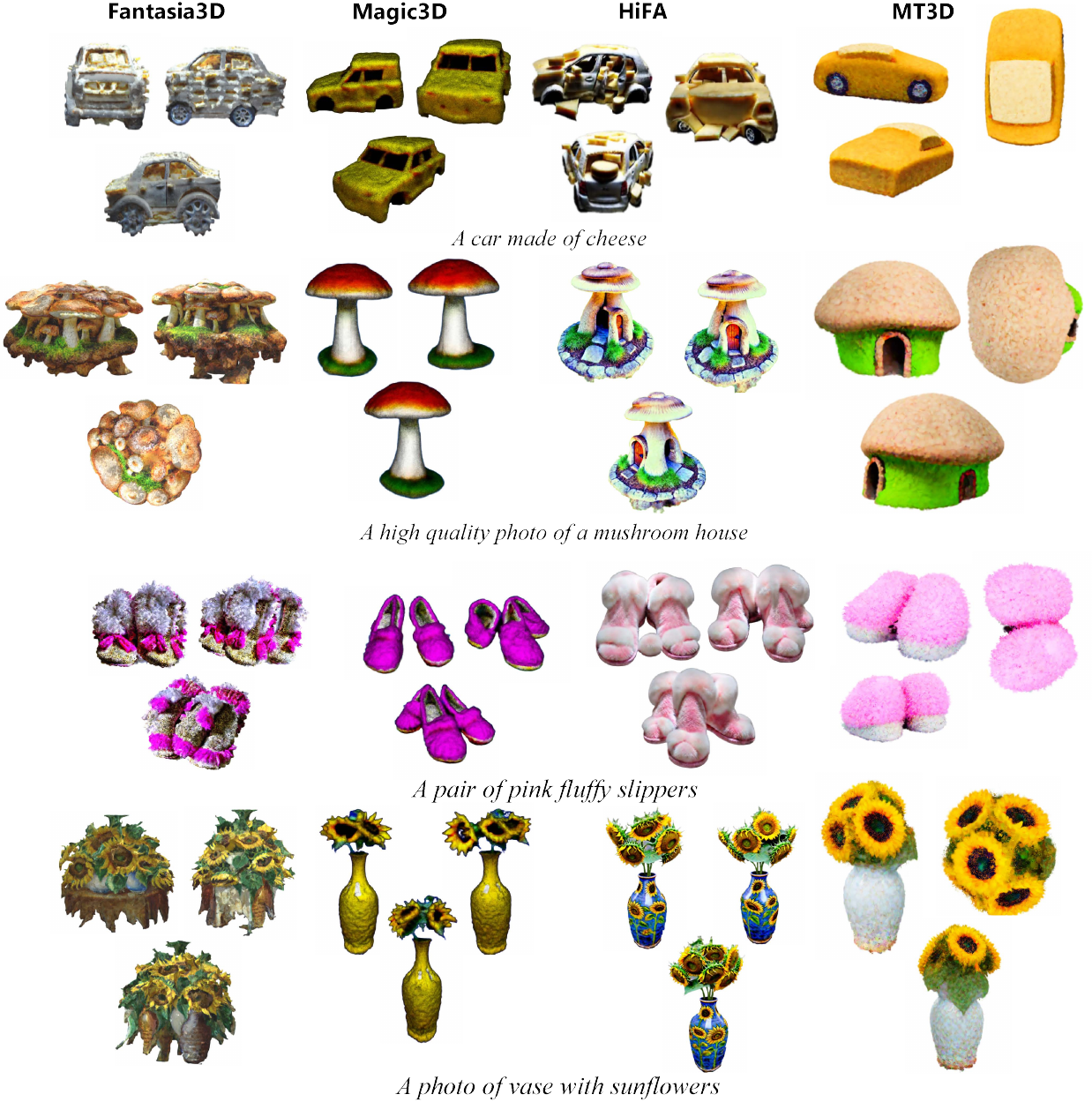}
        \caption{More comparisons between MT3D and state-of-the-art generators}
        \label{fig:Comparison_sota1}
\end{figure*}

\begin{figure*}[h]
    \centering
    \includegraphics[width=0.95\linewidth]{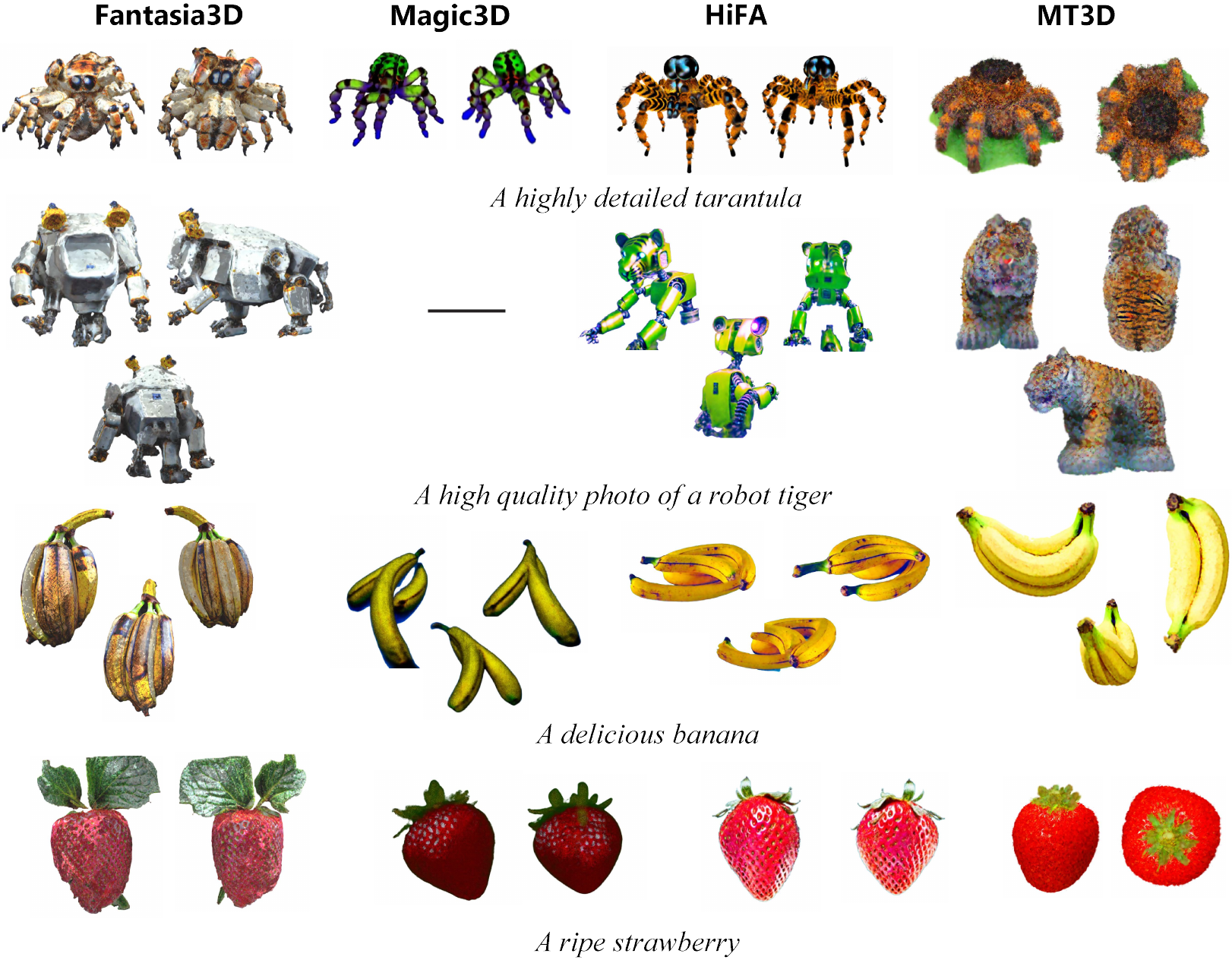}
        \caption{More comparisons between MT3D and state-of-the-art generators}
        \label{fig:Comparison_sota3}
\end{figure*}

In this section we provide additional qualitative comparisons between the proposed MT3D and state-of-the-art generators, including Fantasia3D, Magic3D, and HiFA. Our proposed MT3D generates geometrically consistent renderings in most cases. For instance, for the prompt "A car made of cheese," other state-of-the-art generators fail to maintain the geometric features of the car and instead render pieces of cheese. In contrast, MT3D preserves the car's geometry while applying a cheese texture. Additionally, several other examples in Figures \ref{fig:Comparison_sota2},\ref{fig:Comparison_sota1} and \ref{fig:Comparison_sota3}, further validate the superior performance of our proposed method compared to other state-of-the-art generators.

\subsection{High-fidelity 3D objects used to guide MT3D}
\begin{figure*}[h]
    \centering
    \includegraphics[width=1.0\linewidth]{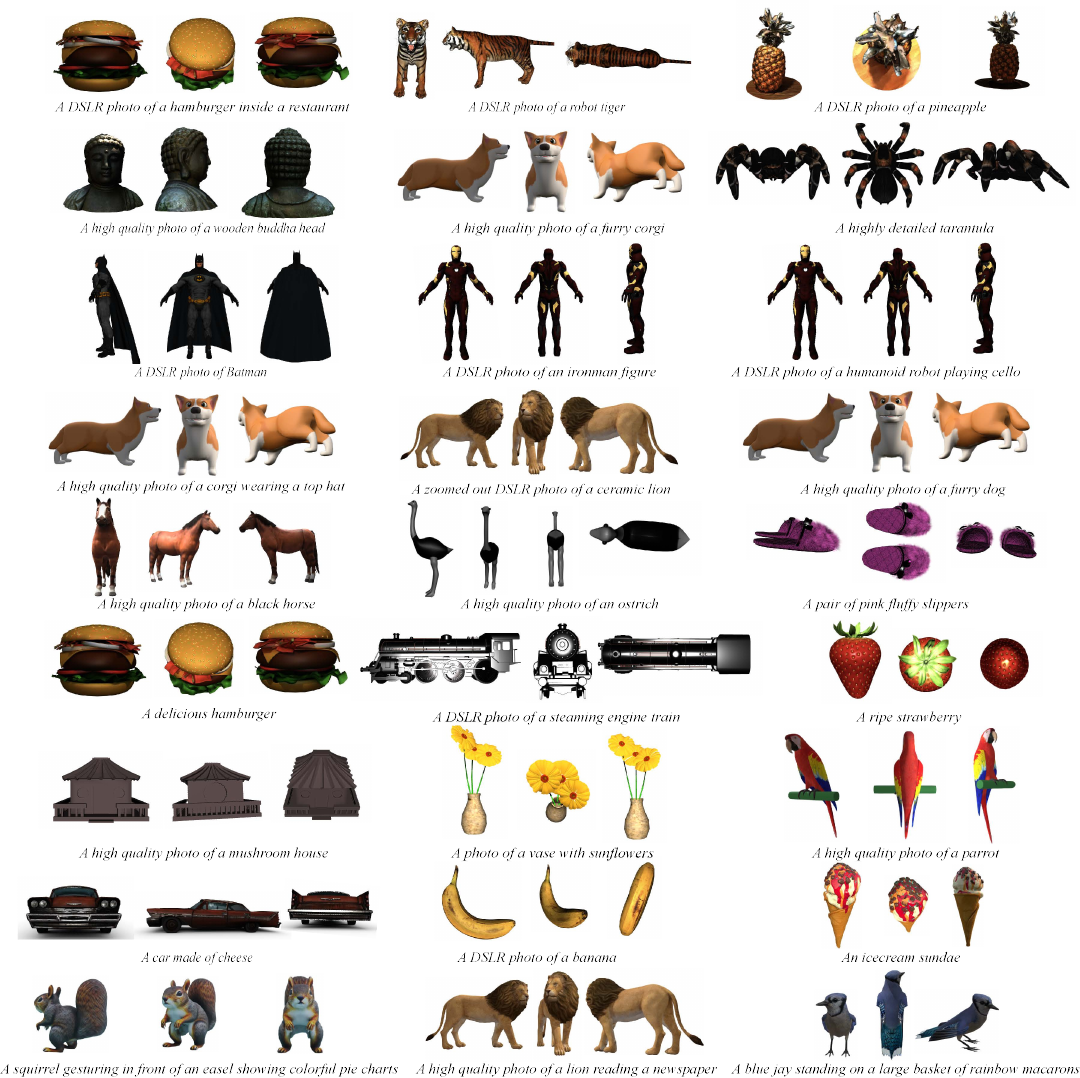}
        \caption{Illustrates high-fidelity 3D objects corresponding to various text prompts used for guiding
MT3D throughout the paper.}
        \label{fig:gt1}
\end{figure*}

In Figures \ref{fig:gt1}, we present the high-fidelity 3D objects from Objaverse used to guide the generation with MT3D.

\end{document}